\documentclass{article}

\usepackage{microtype}
\usepackage{graphicx}
\usepackage{booktabs}
\usepackage{amsmath}
\usepackage{amssymb}
\usepackage{mathtools}
\usepackage{amsthm}
\usepackage{hyperref}

\usepackage[preprint]{icml2026}

\theoremstyle{plain}
\newtheorem{proposition}{Proposition}
\newtheorem{lemma}{Lemma}

\theoremstyle{definition}
\newtheorem{definition}{Definition}
\theoremstyle{remark}

\icmltitlerunning{RSD for Moving Local Audits of Language-Model Hidden States}

\begin{document}

\twocolumn[
\icmltitle{RSD: Moving Local Triangular Charts for Auditing Language-Model Hidden States}

\begin{icmlauthorlist}
\icmlauthor{Seungmin Jin}{hse}
\end{icmlauthorlist}

\icmlaffiliation{hse}{HSE University}
\icmlcorrespondingauthor{Seungmin Jin}{sedzhin@hse.ru}
\icmlkeywords{explainable AI, representation auditing, language models, local chart audit, residual analysis}

\vskip 0.3in
]

\printAffiliationsAndNotice{}

\begin{abstract}
We study Relational Semantic Decomposition, abbreviated as RSD, as a moving local triangular chart audit for language-model hidden states. For repeated occurrences of one target word, RSD fits a shared three-anchor membership chart $S_t$ at layer or token-time $t$. The hidden-state channel uses $X_t\approx S_tC_t$; the invariant readout $M_t=S_tS_t^\top$ is the induced occurrence co-membership relation, and $R_t=X_t-S_tC_t$ records what the fitted root chart leaves outside the chart. The broader joint audit reuses the same membership chart for relation data, $A_t\approx S_tB_tS_t^\top$, such as an attention-derived occurrence relation. The current GPT-2 evidence is the $X$-channel hidden-state audit with Word-in-Context labels used as an external same-sense versus different-sense reference relation. On full WiC train, the root chart passes 16 of 53 eligible target words; this is audit coverage, not GPT-2 task accuracy. Token-time and pair-level diagnostics show the main regimes: \texttt{make} and \texttt{break} align at the target state, \texttt{drive} and \texttt{stay} improve after right context in small-count exploratory cases, and \texttt{play} remains a localized root-chart failure whose final same-sense pairs are not closer and have larger residual discrepancy. The resulting claim is diagnostic: RSD reports where a sense relation is visible in root co-membership and which failures become residual branch candidates or attention-channel obligations.
\end{abstract}

\section{Introduction}

Language-model hidden states carry local relation structure that is not visible from a final task score alone. This paper studies a concrete version of that problem. Given many occurrences of the same target word, we ask where GPT-2 represents the pair relation supplied by Word-in-Context, abbreviated as WiC: two occurrences have either the same sense or different senses. The output is an audit report, not a classifier. It says whether the relation is visible in a root hidden-state chart, whether it appears only after more context, or whether it remains outside the root chart.

Relational Semantic Decomposition, abbreviated as RSD, provides the chart. For a fixed word-conditioned object set at layer or token-time $t$, the hidden-state matrix is
\begin{equation}
X_t\in\mathbb{R}^{n\times d},
\end{equation}
with one row per target-word occurrence. RSD fits a three-anchor membership chart
\begin{equation}
S_t\in\Delta_3^n,
\end{equation}
and anchor coordinates
\begin{equation}
C_t\in\mathbb{R}^{3\times d},
\end{equation}
so that
\begin{equation}
X_t\approx S_tC_t.
\end{equation}
The primary relation readout is
\begin{equation}
M_t=S_tS_t^\top,
\end{equation}
which is invariant to anchor renaming. The residual
\begin{equation}
R_t=X_t-S_tC_t
\end{equation}
records coordinate signal not explained by the root chart.

The running story is deliberately pairwise. If two WiC occurrences of \texttt{make} or \texttt{break} have the same sense, the root chart should make their membership rows close; different-sense pairs should be farther apart. For these words, the target-state audit shows this desired ordering. For \texttt{play}, the same check fails: same-sense pairs are not reliably closer even at the final state, and their residual discrepancy is larger. The failure is therefore localized. It is not presented as GPT-2 task failure. It is a statement that this root chart does not expose the WiC relation in $M_t$, so \texttt{play} becomes a residual branch candidate and an attention-channel test case for the next audit.

WiC is used only as an external reference relation. We do not train a WiC classifier, report GPT-2 WiC accuracy, or claim that WiC labels are the model's internal semantic ground truth. The audit asks a narrower question: does the model-internal co-membership relation $M_t$ agree with the external same-sense versus different-sense relation within the declared object set, and where does disagreement go?

The paper makes three contributions. First, it formulates RSD as a moving local triangular chart audit for language-model hidden states, with $M_t=S_tS_t^\top$ as the permutation-invariant occurrence relation readout. Second, it defines pair-level failure diagnostics using membership distance and residual discrepancy, separating root co-membership failure from residual branch candidates. Third, it reports GPT-2 WiC audits that separate relation regimes: full WiC train gives 16 root-chart pass words out of 53 eligible target words, token-time diagnostics distinguish target-state and final-state alignment, and the word \texttt{play} becomes a concrete persistent failure case.

\section{Method}

\subsection{Word-Conditioned Object Sets}

Each audit starts from a finite object set: all selected occurrences of one target word. The object identities stay fixed while the representation changes across layer or token-time. In the GPT-2 WiC experiments, $t$ is either a GPT-2 layer at the target token or a token-time comparison between the target position and the final position. This makes the audit local. It does not seek a global semantic basis shared across all words.

\subsection{Root Hidden-State Chart}

For each step $t$, RSD fits a triangular fuzzy membership chart with three anchors:
\begin{equation}
S_t\in\mathbb{R}_{\ge0}^{n\times 3},\quad \sum_{k=1}^{3}s_{i,t,k}=1.
\end{equation}
The coordinate channel is reconstructed as
\begin{equation}
\widehat X_t=S_tC_t,
\end{equation}
with residual
\begin{equation}
R_t=X_t-S_tC_t.
\end{equation}
Rows of $S_t$ are fuzzy barycentric memberships. Rows of $C_t$ are local anchor vectors.

The implementation used here uses an orthogonal moving frame. Let
\begin{equation}
\mu_t=\frac{1}{n}\sum_{i=1}^{n}x_{i,t}
\end{equation}
be the local center, let $Q_t\in\mathbb{R}^{2\times d}$ be a local two-dimensional orthonormal frame, and let $U\in\mathbb{R}^{3\times 2}$ be a fixed equilateral triangle. The anchors have the form
\begin{equation}
C_t=\mathbf{1}\mu_t^\top+r_tUQ_t.
\end{equation}
This construction gives a controlled local chart for the selected object set. The anchor names themselves have no semantic status.

\subsection{Co-Membership Readout}

The chart report is the co-membership relation
\begin{equation}
M_t=S_tS_t^\top,
\end{equation}
with entries
\begin{equation}
M_{ij,t}=s_{i,t}^\top s_{j,t}.
\end{equation}
We also use the membership-row distance
\begin{equation}
d_S(i,j,t)=\lVert s_{i,t}-s_{j,t}\rVert_2.
\end{equation}
For a WiC pair, same-sense occurrences should have larger $M_{ij,t}$ or smaller $d_S(i,j,t)$ than different-sense occurrences if the root chart exposes the reference relation. This is the audit criterion. It is not a classifier objective and it is not a claim that $M_t$ is semantic ground truth.

\subsection{Residual and Branch Diagnostics}

Root failures are useful only if they say where to look next. For residual rows $r_{i,t}$, define
\begin{equation}
e_{i,t}=\lVert r_{i,t}\rVert_2^2
\end{equation}
and pair residual discrepancy
\begin{equation}
\delta_R(i,j,t)=\lvert e_{i,t}-e_{j,t}\rvert.
\end{equation}
Large residual discrepancy means that the pair can be close or far in membership space while still differing in coordinate signal outside the fitted root chart. We therefore report such pairs as residual-branch candidates. This is only a triage label: it does not by itself prove that a residual child chart exists. A stronger residual-branch claim must first rule out root-anchor artifacts, for example by checking the fixed-membership projection residual $R_t^{\star}=X_t-S_tS_t^+X_t$, then testing residual stability and held-out relation gain against controls.

\subsection{Relation Channel Obligation}

The full dynamic audit reuses the same membership chart for relation data:
\begin{equation}
A_t\approx S_tB_tS_t^\top.
\end{equation}
Here $A_t$ may be derived from GPT-2 attention, behavior probes, context similarity, or another declared occurrence relation. In this paper, the reported WiC evidence is the $X$-channel hidden-state audit. The equation above states the next obligation: if the root chart misses a relation in $M_t$, first certify the residual branch candidate under projection-residual and stability controls, then test whether the same membership logic explains an attention-derived $A_t$ under shuffled-proxy and capacity controls.

\section{Formal Audit Objects}
\label{sec:formal-audit}

This section states the algebraic objects needed to interpret the audit. The statements are local. They concern one object set, one hidden-state channel, one chart family, and one step or trajectory.

\begin{lemma}
\textbf{Simplex membership.}
Let $a_{ik}>0$ and define
\begin{equation}
s_{ik}=\frac{a_{ik}}{\sum_{r=1}^{K}a_{ir}}.
\end{equation}
Then $s_{ik}\ge0$ and $\sum_{k=1}^{K}s_{ik}=1$ for every item $i$.
\end{lemma}

\textit{Sketch.} Positivity gives nonnegativity. The denominator is the row sum, so summing over $k$ gives one.

\begin{proposition}
\textbf{Anchor-label invariance.}
Let $P$ be a $K\times K$ permutation matrix. Replacing $S$ by $SP$ and $C$ by $P^\top C$ leaves $SC$ and $R=X-SC$ unchanged. The co-membership matrix $M=SS^\top$ is also unchanged.
\end{proposition}

\textit{Sketch.} Since $PP^\top=I$, we have $(SP)(P^\top C)=SC$. Also $(SP)(SP)^\top=SPP^\top S^\top=SS^\top$. Thus $M$ is the natural invariant relation readout for a fuzzy chart whose anchor names are arbitrary.

\begin{definition}
\textbf{Root-chart relation alignment.}
For an external pair relation $Y_{ij}$, a root chart is aligned with that relation when the score $M_{ij}$, or equivalently the distance $d_S(i,j)=\lVert s_i-s_j\rVert_2$, orders same-relation and different-relation pairs in the expected direction within the declared object set.
\end{definition}

This definition is deliberately conditional. It does not say that $M$ is semantic ground truth. It says that $M$ is the model-internal co-membership relation produced by the chart, and it can be compared with an external reference relation such as WiC.

\begin{definition}
\textbf{Residual discrepancy.}
For residual rows $r_i=x_i-s_iC$, define
\begin{equation}
e_i=\lVert r_i\rVert_2^2
\end{equation}
and
\begin{equation}
\delta_R(i,j)=\lvert e_i-e_j\rvert.
\end{equation}
Large $\delta_R$ means that two items may be close or far in membership space while still differing strongly in coordinate signal outside the root chart.
\end{definition}

\begin{definition}
\textbf{Branch candidates.}
A pair is a residual-branch candidate when its residual discrepancy is high relative to the local table. A pair is an anchor-branch candidate when its membership distance or top-anchor sector indicates that the root chart separates the items at the simplex level. These are audit triage labels. They are not semantic class names.
\end{definition}

The formal obligation for recursive RSD is then concrete. Branch labels remain diagnostic candidates until the next audit proves more. For a residual branch, the next audit must use a declared residual object, such as the fixed-membership projection residual $R^{\star}=X-SS^+X$, rule out anchor-coordinate artifacts, and show stable held-out relation gain against controls. Without these checks, the paper should not claim that a residual child chart exists.

\section{Experiments}

The experiments are organized as a diagnostic path. Step 1 asks what the root chart reports across eligible WiC target words. Step 2 asks when the relation appears by comparing target and final token-time states. Step 3 localizes a persistent failure by comparing membership distance and residual discrepancy for selected word pairs. Throughout, WiC labels are reference edges between target-word occurrences, not training labels for a classifier.

\subsection{Positive Controls: Behavior-Linked Polysemy}

We first check whether RSD memberships can align with behavior-linked hidden-state variation in controlled ambiguous-word fixtures. The words \texttt{bank}, \texttt{bat}, and \texttt{seal} are audited against behavior-margin axes derived from GPT-2 continuation preferences. These controls support an existence claim: a local triangular chart can track behavior-relevant variation in small designed settings.

For \texttt{bank}, the natural continuation probe reaches behavior polar accuracy $1.0$. Held-out behavior-margin alignment passes shuffled controls at layers 1 and 10, with held-out correlations $0.998$ and $0.977$. For redesigned \texttt{bat} prompts, behavior accuracy is $1.0$ and layers 1 through 11 beat shuffled controls; the strongest layer is layer 11 with held-out correlation $0.987$. For \texttt{seal}, layers 1 through 11 also beat shuffled controls after correcting the signed animal-artifact axis. These controls are not WiC results and do not establish broad semantic generalization.

\subsection{Step 1: Root-Chart WiC Coverage}

We next audit GPT-2 hidden states on WiC train. Eligible words must have at least 20 examples and at least three same-sense and three different-sense pairs. This yields 53 target words and 1906 pairs. The root chart passes 16 of 53 eligible words at the final layer. The passing words are \texttt{beat}, \texttt{break}, \texttt{cover}, \texttt{face}, \texttt{get}, \texttt{give}, \texttt{go}, \texttt{have}, \texttt{head}, \texttt{line}, \texttt{make}, \texttt{place}, \texttt{pull}, \texttt{set}, \texttt{turn}, and \texttt{work}.

The coverage number is deliberately not the headline. It is not GPT-2 WiC accuracy. It measures how often the root triangular chart makes WiC same-sense pairs closer in membership space than different-sense pairs under the word-conditioned audit. The result says that root-level co-membership explains some occurrence relations and leaves many words for residual, anchor, or relation-channel analysis.

\begin{table}
\caption{Step 1 root-chart WiC audit coverage on GPT-2 hidden states. The unit is an eligible target word, not an individual prediction.}
\label{tab:wic-coverage}
\centering
\small
\setlength{\tabcolsep}{4pt}
\begin{tabular}{l r}
\toprule
Quantity & Value \\
\midrule
Train examples & 5428 \\
Eligible target words & 53 \\
Audited pairs & 1906 \\
Final-layer pass words & 16 \\
Final-layer fail words & 37 \\
Mean final RSD AUC & 0.606 \\
First-token or pre-context fraction & 0.395 \\
\bottomrule
\end{tabular}
\end{table}

\subsection{Step 2: Token-Time Alignment Regimes}

Causal GPT-2 makes the target-position state context-sensitive. If the target word appears before the informative right context, the target state may not yet contain the relation needed by WiC. We therefore compare the target state with the final state for selected words. Table~\ref{tab:token-time} reports the best target-state and final-state AUC over layers, using negative membership distance as the score for same-sense labels.

\begin{table}
\caption{Step 2 token-time RSD audit over selected WiC words. The metric is AUC of negative membership distance for WiC same-sense labels.}
\label{tab:token-time}
\centering
\small
\setlength{\tabcolsep}{3pt}
\begin{tabular}{lrrrr}
\toprule
Word & $n$ & Same & Best target & Best final \\
\midrule
\texttt{make} & 40 & 25 & 0.777 & 0.795 \\
\texttt{break} & 40 & 16 & 0.799 & 0.703 \\
\texttt{cover} & 28 & 4 & 1.000 & 0.792 \\
\texttt{give} & 40 & 10 & 0.878 & 0.627 \\
\texttt{play} & 40 & 24 & 0.574 & 0.469 \\
\texttt{drive} & 27 & 3 & 0.576 & 0.889 \\
\texttt{stay} & 23 & 6 & 0.539 & 0.716 \\
\bottomrule
\end{tabular}
\end{table}

The table separates regimes rather than ranking models. \texttt{make}, \texttt{break}, and \texttt{give} show strong target-state alignment. \texttt{drive} and \texttt{stay} improve at the final state, although their same-sense counts are small and the result should be treated as exploratory. \texttt{play} remains weak at both target and final states. Token-time therefore turns a raw pass or fail outcome into a diagnostic path: relation visible now, relation emerging after context, or relation still missing from the root chart.

\subsection{Step 3: Root-Pair Failure Localization}

We then inspect pair-level diagnostics for \texttt{make}, \texttt{break}, and \texttt{play} at layer 12. Table~\ref{tab:root-pair} reports mean membership distance $d_S$ and residual discrepancy $\delta_R$ for same-sense and different-sense pairs.

\begin{table}
\caption{Step 3 root-chart pair audit at layer 12. Same and different refer to WiC labels. $d_S$ is membership distance. $\delta_R$ is residual discrepancy.}
\label{tab:root-pair}
\centering
\small
\setlength{\tabcolsep}{3pt}
\begin{tabular}{llrrrr}
\toprule
Word & Step & same $d_S$ & diff $d_S$ & same $\delta_R$ & diff $\delta_R$ \\
\midrule
\texttt{make} & target & 0.234 & 0.824 & 618.2 & 750.9 \\
\texttt{make} & final & 0.405 & 0.498 & 374.7 & 840.6 \\
\texttt{break} & target & 0.280 & 0.712 & 3516.4 & 3850.1 \\
\texttt{break} & final & 0.484 & 0.322 & 639.5 & 685.0 \\
\texttt{play} & target & 0.492 & 0.482 & 866.7 & 764.0 \\
\texttt{play} & final & 0.461 & 0.414 & 913.2 & 613.1 \\
\bottomrule
\end{tabular}
\end{table}

For \texttt{make} and \texttt{break}, the target-state $d_S$ values separate same and different pairs in the desired direction. For \texttt{play}, the final same-sense pairs are not closer than different-sense pairs, and their residual discrepancy is larger. This is the central failure story. The root co-membership relation $M_t$ does not expose the WiC relation for \texttt{play}; the residual diagnostic shows that same-sense pairs can still differ in coordinate signal outside the triangular chart.

These experiments support a diagnostic conclusion. RSD reports where a word-conditioned sense relation is visible in the root co-membership relation, where it appears only after additional context, and which failures become residual branch candidates or attention-channel test cases. They do not yet show that a residual child chart recovers the failed relation.

\section{Related Work}

Coordinate decompositions such as PCA and spectral methods explain dominant variance directions in a coordinate matrix \citep{pearson1901lines}. NMF explains a matrix through nonnegative parts-based factors \citep{lee1999learning}. Archetypal analysis represents data through convex combinations of extreme prototypes \citep{cutler1994archetypal}, and CUR decompositions improve interpretability by selecting actual rows or columns \citep{mahoney2009cur}. These methods are natural coordinate baselines for $X$; in their standard form, they do not report shared-membership compatibility with a named weak pairwise proxy $A$.

Graph and joint-factorization models also combine coordinate and relation information. Graph-regularized NMF uses a graph smoothness penalty while factorizing $X$ \citep{cai2011graph}. Mixed-membership stochastic blockmodels explain relational data through node-level memberships \citep{airoldi2008mixed}. Collective matrix factorization shares latent factors across multiple relation matrices \citep{singh2008relational}, and nonnegative tri-factorization connects matrix factorization with clustering structure \citep{ding2006orthogonal}. RSD is narrower than these predictive or clustering models. It treats $A$ as a named weak proxy and fits a local membership geometry under a declared decoder class. It then reports coordinate loss, proxy loss, component mass, and residual outliers as audit quantities.

Fuzzy clustering and relation decomposition methods naturally target pairwise relation or affinity objects \citep{mirkin1996mathematical}. FADDIS and additive fuzzy spectral clustering are relation-factor methods for similarity or affinity data \citep{mirkin2012additive}. They target relation or affinity objects such as $A$. RSD differs by asking for a shared-membership cross-view witness between $A$ and $X$, then inspecting the empirical residual $R=X-SC$.

Lexical resources and word embeddings supply useful reference relations and compact controls. WordNet provides controlled lexical relations \citep{miller1995wordnet}; GloVe provides static coordinate vectors \citep{pennington2014glove}. Recent work links calendar-like representational manifolds to symmetry in language statistics \citep{karkada2026symmetry}. In this paper, the main evidence is instead a GPT-2 hidden-state audit over WiC target-word occurrences. Static lexical fixtures remain useful as sanity checks, but they are not the central claim.

Probing and representation analysis ask whether model states contain recoverable information \citep{alain2016understanding,belinkov2019analysis,hewitt2019structural,tenney2019bert}. RSD asks a different local question. It asks whether a word-conditioned hidden-state cloud admits a triangular membership geometry whose co-membership agrees with an external occurrence relation, and it reports residual structure when the root chart fails.

\section{Limitations}

The current GPT-2 WiC evidence is a hidden-state audit, not a supervised benchmark result. No WiC classifier is trained, no GPT-2 task accuracy is reported, and the AUC values are diagnostics of membership-distance alignment with an external reference relation. They should not be read as downstream task performance.

The reported WiC tables are $X$-channel root-chart audits. The broader RSD model requires the same membership chart to explain relation data through $A_t\approx S_tB_tS_t^\top$, but attention-derived $A_t$ is not yet integrated into the empirical WiC results. Therefore the evidence supports root-chart hidden-state diagnostics and failure localization, not a completed joint hidden-state and attention-channel decomposition.

The branch labels remain triage labels. A residual-branch candidate is a pair or subset with large residual discrepancy under the root chart. This is weaker than a residual child-chart claim. To support such a claim, the next audit must compute a declared projection residual such as $R^{\star}=X-SS^+X$, rule out root-anchor artifacts, check stability, and show held-out relation gain against controls. The present GPT-2 WiC evidence has not completed those checks.

The word-conditioned WiC analysis uses local subsets with uneven label counts. Some token-time results, especially for \texttt{cover}, \texttt{drive}, and \texttt{stay}, have small same-sense counts. They are useful for failure taxonomy, but they do not support broad generalization claims. Stronger claims about language-model semantics would require more models, more datasets, seed and subsample stability, and baselines such as cosine distance, supervised probes, and graph-factorization controls.

Finally, WiC labels are an external semantic reference, not ground truth about GPT-2's internal organization. A mismatch can mean that the chart is inadequate, that the hidden states encode another relation, that the benchmark relation is too coarse for this local object set, or that tokenization and causal context create representation states not captured by the current audit.

\section{Conclusion}

RSD gives a moving local triangular chart for auditing language-model hidden states. The central readout is the co-membership relation $M_t=S_tS_t^\top$, supported by the residual $R_t=X_t-S_tC_t$. This lets the audit ask where a word-conditioned sense relation is visible in the root chart and where the chart fails.

The GPT-2 WiC audits show three regimes. \texttt{make} and \texttt{break} have target-state co-membership aligned with WiC same-sense relations. \texttt{drive} and \texttt{stay} improve at the final state in small-count exploratory cases, showing why token-time matters. \texttt{play} remains the persistent failure: final same-sense pairs are not closer in membership space, and their residual discrepancy is larger than for different-sense pairs.

The contribution is therefore a relation-audit interpretation. Root-chart coverage, token-time dynamics, and pair-level residual discrepancy identify where the relation is visible and which failures become branch candidates. If the root chart misses a relation, the next step is not to assume a child chart. It is to certify the residual branch candidate with projection-residual, stability, and held-out controls, and to test GPT-2 attention through an explicit relation channel $A_t\approx S_tB_tS_t^\top$.

\bibliography{custom}
\bibliographystyle{icml2026}

\end{document}